\documentclass[letterpaper,10pt,conference]{ieeeconf}
\IEEEoverridecommandlockouts
\makeatletter
\let\NAT@parse\undefined
\makeatother
\usepackage{algorithm}
\usepackage[noend]{algpseudocode}
\usepackage{amsfonts}
\usepackage{amsmath}
\usepackage{amssymb}
\usepackage{array}
\usepackage{bm}
\usepackage{color}
\usepackage{comment}
\usepackage{float}
\usepackage{gensymb}
\usepackage{graphicx}
\usepackage[colorlinks]{hyperref}
\usepackage{multirow}
\usepackage{nicefrac}
\usepackage[caption=false,font=footnotesize,subrefformat=parens,labelformat=parens]{subfig}
\usepackage{textcomp}
\usepackage{wrapfig}

\graphicspath{{figures/}{images/}}

\newcommand\Set[2]{\{\,#1\mid#2\,\}}

\title{An Uncertainty Estimation Framework for Probabilistic Object Detection}

\author{Zongyao Lyu, Nolan B. Gutierrez, and William J. Beksi
\thanks{Z. Lyu, N.B. Gutierrez, and W.J. Beksi are with the Department of Computer
        Science and Engineering, University of Texas at Arlington, Arlington,
        TX, USA.
        Emails:
        zongyao.lyu@mavs.uta.edu,
        nolan.gutierrez@mavs.uta.edu,
        william.beksi@uta.edu
        }
}

\begin{document}
\maketitle
\pagestyle{plain}

\begin{abstract}
In this paper, we introduce a new technique that combines two popular methods to
estimate uncertainty in object detection. Quantifying uncertainty is critical in
real-world robotic applications. Traditional detection models can be ambiguous
even when they provide a high-probability output. Robot actions based on
high-confidence, yet unreliable predictions, may result in serious
repercussions. Our framework employs deep ensembles and Monte Carlo dropout for
approximating predictive uncertainty, and it improves upon the uncertainty
estimation quality of the baseline method. The proposed approach is evaluated on
publicly available synthetic image datasets captured from sequences of video.
\end{abstract}


\section{Introduction}
\label{sec:introduction}
Object detection is a central task in computer vision and a necessary capability
for many robotic operations. Models for object detection have achieved high mean
average precision (mAP) on datasets such as ImageNet
\cite{russakovsky2015imagenet}, the PASCAL Visual Object Classes
\cite{everingham2010pascal,everingham2015pascal}, and the Microsoft Common
Objects in Context (COCO) \cite{lin2014microsoft}. However, these models can
fail when evaluated in dynamic scenarios which commonly contain objects from
outside the training dataset. The mAP metric encourages detectors to output many
detections for each image, yet it does not provide a consistent measure of
confidence other than a higher score indicating that the system is more
accurate. Nonetheless, robotic vision systems must be able to cope with diverse
operating conditions such as variations in illumination and occlusions in the
surrounding environment.


The ACRV Robotic Vision Challenge on Probabilistic Object Detection (PrOD)
\cite{skinner2019probabilistic} introduces a variation on conventional object
detection tasks, which requires quantification of spatial and semantic
uncertainty with the use of a new evaluation measure called the
probability-based detection quality (PDQ) \cite{hall2020probabilistic}.
Different from mAP, the PDQ jointly evaluates the quality of spatial and label
uncertainty, foreground and background separation, and the number of true
positive (correct), false positive (spurious), and false negative (missed)
detections. In addition, the datasets provided by the challenge are video
sequences comprising various scenes that have been generated from high-fidelity
simulations.

\begin{figure}[ht]
\vspace{-2mm}
\centering
\subfloat[]{
  \includegraphics[width=0.31\columnwidth]{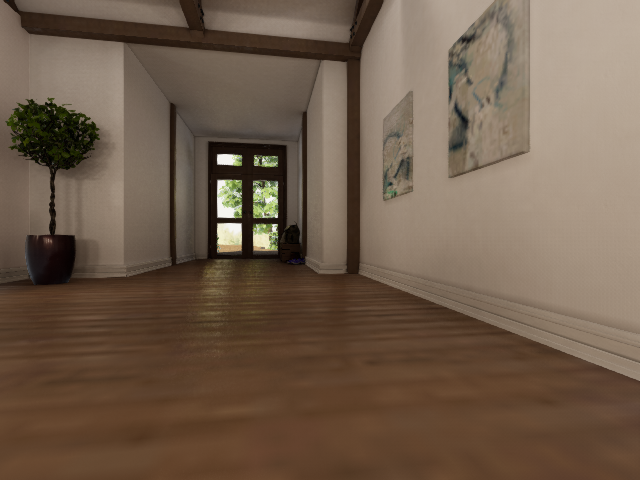}
}
\subfloat[]{
  \includegraphics[width=0.31\columnwidth]{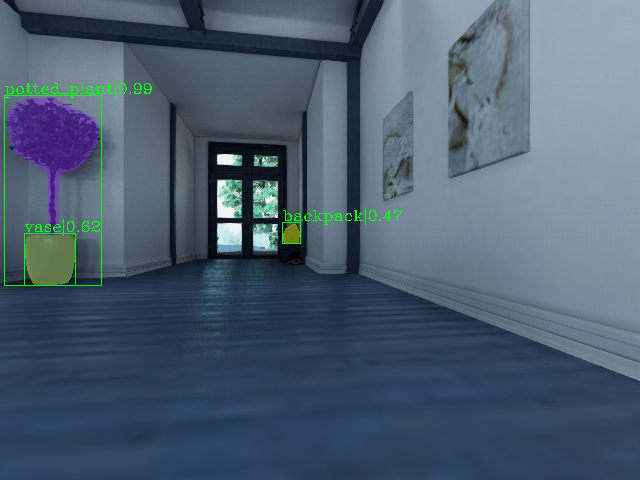}
}
\subfloat[]{
  \includegraphics[width=0.31\columnwidth]{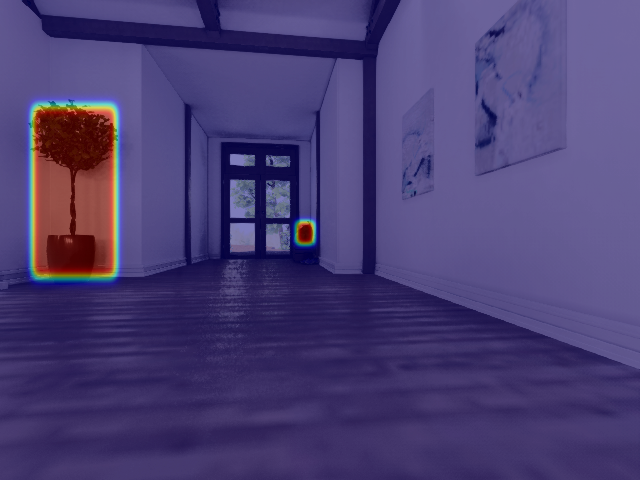}
}
\caption{An example of a raw input image (a), object detections (b), and the
corresponding heatmaps associated with the bounding boxes (c). Our object
detection system represents the location of objects as probabilistic bounding
boxes. This allows a detector to express spatial uncertainty by inducing a
probability distribution over the pixels of the object.}
\label{fig:probablistic_bounding_boxes}
\vspace{-4mm}
\end{figure}

In this work, we extend our initial groundwork \cite{lyu2020probabilistic} for
addressing the issue of uncertainty estimation in object detection
(Fig.~\ref{fig:probablistic_bounding_boxes}). This includes the exploration of
two popular uncertainty estimation techniques: deep ensembles
\cite{lakshminarayanan2017simple} and Monte Carlo (MC) dropout
\cite{gal2016dropout}. We apply these techniques jointly on two detectors,
Hybrid Task Cascade \cite{chen2019hybrid} and Grid R-CNN \cite{lu2019grid}, to
produce probabilistic bounding boxes which allow a detector to express spatial
uncertainty. Furthermore, semantic (i.e., label) uncertainty is quantified by a
probability distribution over the known classes for each detection. In summary,
our contributions are the following.
\begin{itemize}
  \setlength{\itemsep}{-\parsep}\setlength{\topsep}{-\parsep}
  \item We devise a novel aggregation of two prominent techniques to quantify
  uncertainty in object detection models.
  \item We create a new heuristic for improving false positive label quality.
  \item We provide an easy-to-implement method for merging ensembles of
  detections.
  \item We introduce high-speed tuning of the hyperparameters for increasing PDQ
  scores.
  \item We show that non-maximum suppression (NMS) can be used as a
  post-processing step to keep all detections.
\end{itemize}
Our source code is available at \cite{depod}.

The remainder of this paper is organized as follows. We describe related work in
Section~\ref{sec:related_work}. Section~\ref{sec:background} provides the
background information for calculating the PDQ score. Our framework for
probabilistic object detection is presented in
Section~\ref{sec:probabilistic_object_detection_framework}. In
Section~\ref{sec:experimental_results}, we report our experimental results on
the validation data provided by the PrOD challenge. We conclude in
Section~\ref{sec:conclusion_and_future_work} with a discussion of future work.

\section{Related Work}
\label{sec:related_work}
\renewcommand\thesubsection{\Alph{subsection}}
\subsection{Object Detection}
The goal of object detection is to localize and recognize instances of
particular categories of objects within an image. Current state-of-the-art
object detection systems utilize the power of deep neural networks (DNNs)
\cite{szegedy2013deep} to output a softmax score indicating the class label
and a bounding box denoting the location of each object of a known class.
DNN-based object detectors can be categorized into one-stage detection models
such as YOLO \cite{redmon2016you} and its variations
\cite{liu2016ssd,redmon2017yolo9000, redmon2018yolov3,bochkovskiy2020yolov4},
and two-stage detection models such as R-CNN \cite{girshick2014rich} and its
variants \cite{girshick2015fast,ren2015faster,he2017mask,cai2018cascade}.
Both categories have advantages over each other.

Two-stage detectors include a region proposal generator that produces a set of
proposals and extracts features from each proposal in the first stage, followed
by a second stage of classification and bounding-box regression. Conversely,
one-stage detectors conduct the prediction of object classes and bounding boxes
directly on the image feature maps without a region proposal generation step.
Two-stage detectors generally achieve higher accuracy, but they have a
relatively slow detection speed. On the other hand, one-stage detectors are more
time efficient and have greater applicability to real-time detection,
nevertheless they exhibit lower accuracy.

Although the performance of object detectors has been greatly improved with
the recent development of deep learning algorithms, current state-of-the-art
DNN-based detectors may give high classification probabilities to falsely
detected objects. This occurs due to the fact that they output deterministic
predictions of bounding boxes without information regarding how certain the
predicted results are. Thus, DNN detectors can be unreliable in practical
applications and relying on softmax scores and predicted boxes alone is not
sufficient.

\subsection{Bayesian Neural Networks for Uncertainty Estimation}
To cope with these limitations, one may extend the classic neural network to
produce additional data which can be adapted to supplement the following values:
label and spatial uncertainty. This information can then be used to augment the
classification score and regressed bounding boxes, thus producing more reliable
predictions. Traditionally, Bayesian methods automatically infer hyperparameters
by marginalizing them out of the posterior distribution. They can naturally
express uncertainty in parameter estimates and propagate it to the predicted
results.

Bayesian neural networks (BNNs)
\cite{denker1991transforming,mackay1992practical,hernandez2015probabilistic}
offer a natural way to integrate Bayesian modeling with DNNs. BNNs provide a
mathematically grounded foundation to reason about uncertainty in DNNs by
learning a prior distribution over the weights of a neural network. Not only is
each weight in a BNN a random variable, but also the output of the network is a
random variable thereby providing a measure of uncertainty. However, compared to
non-Bayesian neural networks, BNNs are often computationally intractable which
makes them hard to apply in practice.

\subsection{Dropout as a Bayesian Approximation}
Dropout \cite{hinton2012improving,srivastava2014dropout} was first proposed as
a technique for training neural networks by randomly omitting neuron units
during training to prevent overfitting. It helps prevent co-adaptations whereby
a unit relies on several other specific hidden units being present. Dropout also
provides a way of approximately combining many disparate neural network models
efficiently. Maeda et al. \cite{maeda2014bayesian} provide an explanation of
dropout from a Bayesian standpoint. From this perspective, we can view dropout
as a way to deal with the model selection problem by Bayesian model averaging
where each model is weighted in accordance with the posterior distribution.

Gal et al. \cite{gal2015bayesian,gal2016dropout} extend the idea of dropout as
approximate Bayesian inference in deep Gaussian processes. They show that
dropout can be used at test time to impose a Bernoulli distribution over the
network's weights, thus requiring no additional model parameters. They refer to
this technique as MC dropout. Following this idea, Miller et al.
\cite{miller2018dropout,miller2019evaluating} estimate spatial and semantic
uncertainties of detections in open-set conditions as a rejection criteria using
MC dropout. Kraus et al. \cite{kraus2019uncertainty} incorporate the methods
provided by \cite{gal2015bayesian} into an object detector to estimate
uncertainty, while improving the detection performance of the baseline approach.

\subsection{Uncertainty Estimation using Deep Ensembles}
Ensembles are one way of improving accuracy in object detection by incorporating
detection models. The basic idea is that detectors with contrasting network
structures have a different detection performance for distinct objects in an
image. As a result, the accuracy and robustness of the detector predictions can
be improved due to the application of ensemble methods. For example,
Lakshminarayanan et al. \cite{lakshminarayanan2017simple} propose a non-Bayesian
alternative to BNNs by training an ensemble of multiple networks independently.
When combined together, the networks behave differently for given inputs hence
expressing uncertainty.


\subsection{Probabilistic Object Detection}
As opposed to assuming constant uncertainty, a detection system can assess its
own quality by estimating the ambiguity of its task performance.  Kendall and
Gal \cite{kendall2017multitask} identify evaluating spatial and semantic
uncertainties as paramount importance, especially in applications where
classifications need to be safely clarified. Sirinukunwattana et al.
\cite{sirinukunwattana2016locality} regress probability values for the
classification of nuclei in routine colon cancer histology images. Tanno et al.
\cite{tanno2017bayesian} apply Bayesian image quality transfer to dMRI
super-resolution by applying sub-pixel convolutions to 3D tractography. In
addition, the authors provide uncertainty maps on a pixel-wise basis.  A method
for improving detection accuracy by modeling bounding boxes with Gaussian
parameters is put forth by Choi et al. \cite{choi2019aaussianya}. Wang et al.
\cite{wang2019augpod} propose optimizing PDQ scores by focusing on
training-time augmentations. A key contribution of their work was the insight
that a system could become more certain of its predictions by feeding in gamma
corrected images during training time. Ammirato and Berg
\cite{ammirato2019mask} examine how a system designed for mAP, specifically
Mask-RCNN \cite{he2017mask}, performs in an environment that evaluates the PDQ
metric.

In contrast to the aforementioned works, we combine MC dropout and deep
ensembles, and apply it to the object detection problem. We show that the
uncertainty estimation performance can be improved over simply using deep
ensembles. In addition, whereas previous approaches merged observations by
averaging, our approach takes the merged box to be the box for which the system
is more confident. When dealing with ensembles, this approach makes sense since
object detectors are often distinctly confident with predictions of classes and
objects with certain characteristics such as appearance or frequency in the
dataset. 


\section{Background}
\label{sec:background}
Consider a ground-truth object $G$ and a detection $D$. Associated with $G$ is a
segmentation mask $S$ and a label $c$, along with a bounding box $B$. The
segmentation mask $S$ corresponds to the true bounding box and determines which
pixels belong to $G$, while $B$ is the set of pixels within the bounding box
associated with the object. $D$ consists of a probability distribution over a
label vector $\bm{l}$, a segmentation mask $\hat{S}$, and a probability that
some pixel belongs to the segmentation mask $P(\bm{x} \in \hat{S})$. $\hat{S}$
coincides to the pixels within the detection's predicted bounding box. The
spatial quality $Q_{\hat{S}}(G,D)$ of this detection is defined by
\begin{equation}
  exp(\frac{1}{N} \sum\limits_{\bm{x} \in S} log(P(\bm{x} \in \hat{S})) +
   \frac{1}{N}\hspace{-3mm}\sum\limits_{\bm{x} \in \{S - B\}} log(1 - P(\bm{x} \in
   \hat{S}))),
\end{equation}
where $N$ is the number of pixels in the true bounding box $S$, and the two
terms in the exponent are known as the foreground and background losses,
respectively. $Q_S$ is one if the pixels in the ground-truth segmentation are
each given probability one, and all the remaining pixels in the set $S - B$ are
assigned probability zero. Next, we can calculate the label quality $Q_{\bm{l}}(G,D)$
by 
\begin{equation}
  Q_L(G,D) = \bm{l}_c,
\end{equation}
where $\bm{l}_c$ is the probability of the detection belonging to class $c$ as
predicted by the object detector. The pairwise probabilistic detection quality
score (pPDQ) can then be calculated by
\begin{equation}
  \text{pPDQ}(G,D) = \sqrt{Q_{\hat{S}} \cdot Q_L}.
\end{equation}
Let $\bm{q}_{ij}$ represent the pPDQ of the $j$th detection in the $i$th frame.
Then the overall PDQ score can be found by
\begin{equation}
  \text{PDQ}(G,D) = \frac{1}{\sum\limits_{i = 1}^{F} (\text{TP}_i + \text{FN}_i
  + \text{FP}_i)} \sum\limits_{i = 1}^{F}\sum\limits_{j = 1}^{\text{TP}_i}
  \bm{q}_{ij},
\end{equation}
where $F$ constitutes the number of frames, and $\text{TP}_i, \text{FN}_i,
\text{FP}_i$ represent the number of true positives, false negatives, and false
positives in the $i$th frame, respectively. The PDQ measure allows for
uncertainty to be modeled in the location of an object which is represented as a
probabilistic bounding box. In particular, the corners of the bounding box are
2D Gaussians and a covariance matrix is given for each corner to express a
spatial uncertainty over the pixels. The label uncertainty is described as a
full probability distribution over the known classes for each detection.

\section{Probabilistic Object Detection Framework}
\label{sec:probabilistic_object_detection_framework}
\begin{figure*}
\centering
\includegraphics[scale=0.5]{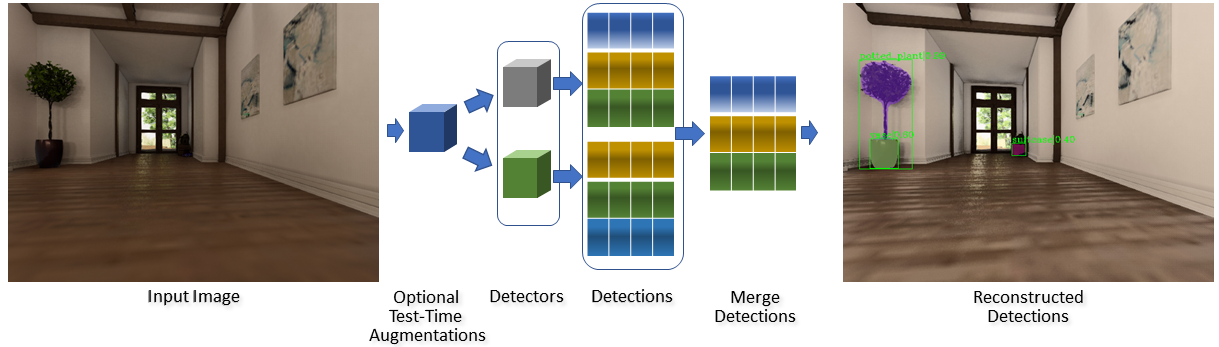}
\caption{An overview of our deep ensemble architecture for probabilistic object
detection.}
\label{fig:model}
\end{figure*}
We detail the key components of our framework for probabilistic object detection
in this section. To begin, we describe the datasets and data preprocessing
steps. Next, we introduce the detection models and explain how our system
utilizes deep ensembles with MC dropout for uncertainty estimation. This is
followed by a discussion of the heuristics, merging methods, and post-processing
techniques we implemented to improve the PDQ scores. An overview of our system
is shown in Fig.~\ref{fig:model}.

\subsection{Datasets and Data Preprocessing}
\label{subsec:data_preprocessing}
The video sequences provided by the PrOD challenge are divided into test,
test-dev, and validation sets. The validation set contains the released ground
truth and is used to evaluate our framework. It consists of around 21500 images
with 56580 ground-truth objects in total. The 30 classes of the validation
dataset are a subset of 80 classes annotated from COCO. We use models pretrained
on COCO and filter out the 50 classes that are not evaluated by the challenge.



\subsection{Detection Models}
\label{subsec:detection_models}
We employ two recent well-performing detection models for the task of
first-stage detection. The first model is Grid-RCNN \cite{lu2019grid}, which
applies a grid-guided mechanism in place of the traditional box offset
regression, for high-quality localization. The second model is Hybrid Task
Cascade (HTC) \cite{chen2019hybrid}, an effective cascade architecture for
object detection and instance segmentation.

\subsection{Combining Deep Ensembles with Monte Carlo Dropout}
\label{subsec:combining_deep_ensembles_with_monte_carlo_dropout}
For each detector, we retain dropout layers in the model during test time and
then make an ensemble of the two models. Our approach is similar to NMS
\cite{li2019teamgl}. We simplify and improve the method proposed by  Miller et
al. \cite{miller2018dropout} in which detections  are grouped into observations
by IoU. A key difference in our approach is we take the merged bounding box from
each group to be the box that the model is most confident about (where
confidence is measured by the highest label score). To merge predictions, we
group detections by IoU via NMS, and we keep the most confident detections from
each observation. 

First, NMS partitions boxes $B$ into groups $G$ such that any two boxes in a
group $G_i$ has an IoU above a threshold $\lambda$ between 0 and 1. Each group
must be sorted by confidence scores. The list of boxes contains the top-left and
bottom-right corners, and the confidence of each box's label. Next, each of
these observations represented as groups of bounding boxes are reduced to the
most confident bounding box and label. Finally, the covariance matrices of every
detection are computed. The procedure for merging the detections from our deep
ensemble is shown in Algorithm~\ref{alg:merge_detections}.


\renewcommand{\algorithmicensure}{\textbf{Output:}}

\begin{algorithm}
\caption{Merging of Deep Ensembles}
\begin{algorithmic}[1]
  \Require {\small I = input image}  
  \Require {\small $D =$ $d_1,d_2,\ldots,d_k$ \Comment{$d_i$: $i^{th}$ detector}}
  \Require {\small $\lambda = [0,1]$ \Comment{IoU between boxes in a cluster}}
  \Require {\small $d_i(I)_B, d_i(I)_L$ \Comment{Detected bounding boxes and labels}}
  \Ensure {$B^{new},L^{new}$ \Comment{\small Merged list of bounding boxes, labels}}
  \State $ B :=  \underset{d \in D}{||} d(I)_B, L :=  \underset{d \in D}{||}
d(I)_L$ \Comment{$||:$ Concatenation}
  \State{G $:= \Set{g}{\forall x,y \in g, \text{IoU}(B_x,B_y) \geq \lambda}$}
  \For{$i$ in $1, 2, 3, \ldots, |G|$}
  \State Sort $G_i$ by scores in $L$ in descending order   
  \State $B_{i}^{new} = B_{G_{i,0}}$
  \State $L_{i}^{new} = L_{G_{i,0}}$
  \EndFor
\end{algorithmic}
\label{alg:merge_detections}
\end{algorithm}

\subsection{Heuristics for Improving the PDQ Score}
\label{subsec:heuristics_for_improving_the_pdq_score}
We can maximize the PDQ score by optimizing metrics that evaluate the quality of
object detections. As an example, we may increase the PDQ score by reducing the
number of objects in a detection that are incorrectly classified
\cite{ammirato2019mask}. Alternatively, researchers have demonstrated that
inflating label scores to one increases the average label quality and thus the
PDQ score. To systematically find the optimal parameters, we save the boxes that
are obtained from each configuration of the model and uniformly search the
hyperparameter space. In the following paragraphs, we describe the details for
each heuristic employed to improve the PDQ score.

{\bf Threshold Filtering:} Rejection of bounding boxes that have low-confidence
has the greatest impact on localization performance and the PDQ score.
Succinctly, the number of false negatives, false positives, and true positives
is affected, therefore producing a wide range of PDQ scores. While other work on
this topic has found an optimal confidence threshold of 0.5
\cite{ammirato2019mask}, we've discovered that higher thresholds significantly
increase the number of false negatives and thus reduce the PDQ score. In our
search of the choice hyperparameters, we've found that we can set the confidence
threshold as low as possible but only when running NMS once more with a low IoU
threshold.

{\bf Bounding Box Reduction:} The PDQ score can be substantially lowered by
confidently labeling pixels in the background. To lessen this issue, we center
crop the bounding box whereby the box's width and height are reduced by a
specified proportion. The PDQ score increases by reducing the ratio of
background pixels to foreground pixels. We specify our reduction ratio to be 0.1
thereby gaining better results.

{\bf Improving False Positive Quality:} False positive label quality is lower
when incorrectly classified objects assign low probabilities to correct labels.
We conceive a novel heuristic by noting that uniformly distributing
probabilities over the rest of the labels in the label vector $l$ can maximize
this metric over all testing data. Concretely,
\begin{equation}
  l_i= 
\begin{cases}
  S,& l_i \text{ corresponds to the detection label,}\\
  \frac{1-S}{30}, & \text{otherwise,}
\end{cases}
\end{equation}
where $S$ is the score of the bounding box's label and 30 is the number of
classes.

{\bf Covariance Matrix Calculation:} We experimented with traditional covariance
calculations and fixed covariances. Notwithstanding, we improve our results by
setting the diagonal of the covariance matrix, which represents the spatial
uncertainty, proportional to the bounding box's width and height
\cite{ammirato2019mask}. As an example, if we use a proportion of $0.10$ for a
50 by 60 bounding box, then the covariance matrix would be defined as
$[[5,0],[0,6]]$ for both the top-left and bottom-right corners. 

{\bf Non-Maximal Suppression:} Contemporary work has shown that the PDQ score
can be improved by removing low-confidence detections. We've verified that this
heuristic does improve the score, nevertheless this result can be outperformed
by adding an additional NMS post-processing step. However, allowing the
evaluation of low-confidence boxes may increase the number of false positives.
To avoid this situation, we allow larger higher-confident boxes to absorb
smaller lower-confident boxes which reduces the number of false positives.

\section{Experimental Results}
\label{sec:experimental_results}
This section presents our experimental results on the validation data provided
by the PrOD challenge. Our implementation is based on MMDetection
\cite{chen2019mmdetection}, an open-source object detection toolbox. The
experiments were performed on a Ubuntu 18.04 machine with an Intel Core i7-8700
CPU, 32GB of RAM, and an NVIDIA Quadro P4000 GPU.

Both detectors used in the experiments are models pretrained on the COCO dataset
with the ResNeXt101 backbone. The results of using an ensemble model composed of
HTC and Grid R-CNN with various threshold parameters are shown in
Table~\ref{tab:validation_set_results_1_threshold}. In our experiments, lower
threshold values provided higher scores.
Table~\ref{tab:validation_set_results_2_boxratio} shows the effect of reducing
the bounding box size of all detections on the PDQ scores. Here we found that
reduction of the bounding box size by 10\% results in better performance.
Table~\ref{tab:validation_set_results_3_covariance} shows how the PDQ score
changes when scaling the covariance to a different percentage of the box size.
We experimented with covariance matrices dependent on the widths and heights of
the bounding boxes. The effectiveness of this technique was demonstrated in
\cite{ammirato2019mask}, and our final results applied this heuristic. After
exploring various values in our implementation, we found that setting the
top-left and bottom-right elements of the covariance matrix equal to a scale of
30\% of the bounding box size gave the best results. We report these findings
with different IoU threshold values in
Table~\ref{tab:validation_set_results_4_iouthreshold}.

Table~\ref{tab:validation_set_results_5} shows the efficacy of the individual
HTC and Grid R-CNN detectors, and their ensembles.
We observed that a single HTC detector outperforms an ensemble of HTC and Grid
R-CNN in our experiments. This indicates that a careful selection of integrated
detection models and parameter tuning are necessary to bring about an effective
ensemble model, thus further exploration is required. We examined MC dropout as
the uncertainty estimation technique on the HTC and Grid R-CNN detectors
individually, and also their ensembles. In practice, this method is equivalent
to performing several stochastic forward passes through the network and then
taking an average of the results. We chose to sample three passes by adding a
dropout rate of 0.3 to the second shared fully-convolutional layer of HTC's
region of interest head.
After testing a large number of post-processing parameters, we found that models
using dropout consistently demonstrated better quality when compared to models
without dropout. The result of an ensemble model with dropout outperforming an
ensemble without dropout demonstrates the effectiveness of our idea for
integrating these two approaches.
The results of the models with dropout are displayed in
Table~\ref{tab:validation_set_results_5}.

Finally, we compared our most confident box merging (MCBM) strategy with
averaging in Table~\ref{tab:validation_set_results_6}. With all other
parameters being the same, we present evidence that even when boxes in an
observation must have the same label, the detection quality is best whenever we
take the merged box to be the most confident box in the observation. Our
hypothesis is we should allow the detection system that is most confident about
its result have its prediction take over the observation. As expected, the MCBM
method achieved the same results with and without the constraint that the boxes
in an observation must have the same label. 

\begin{table*}
\centering
\begin{tabular}{|c|c|c|c|c|c|c|c|c|}
\hline
{\bf Threshold} & {\bf PDQ Score} & {\bf Avg. pPDQ} & {\bf Avg. FP} & {\bf Avg. SQ} & {\bf Avg. LQ} & \bf{TPs} & {\bf FPs} & {\bf FNs} \\
\hline
\hline
\textbf{0.018} & \textbf{22.569} & 0.409 & \textbf{0.835} & 0.464 & 0.498 & \textbf{58961} & 112689 & \textbf{29128} \\
\hline
0.05 & 22.534 & 0.446 & 0.747 & 0.483 & 0.551 & 53036 & 66395 & 35053 \\
\hline
0.1 & 22.468 & 0.481 & 0.648 & 0.498 & 0.602 & 48176 & 42652 & 39913 \\
\hline
0.3 & 22.006 & 0.544 & 0.424 & 0.520 & 0.703 & 39940 & 18615 & 48149 \\
\hline
0.5 & 20.919 & \textbf{0.591} & 0.270 & \textbf{0.537} & \textbf{0.780} & 33809 & \textbf{10277} & 54280 \\
\hline
\end{tabular}
\caption{The results of an ensemble model using different threshold values on
the validation dataset (Avg. SQ=average spatial quality, Avg. LQ=average label
quality, TPs=true positives, FPs=false positives, FNs=false negatives).}
\label{tab:validation_set_results_1_threshold}
\end{table*}

\begin{table*}
\centering
\begin{tabular}{|c|c|c|c|c|c|c|c|c|}
\hline
{\bf Box Ratio} & {\bf PDQ Score} & {\bf Avg. pPDQ} & {\bf Avg. FP} & {\bf Avg. SQ} & {\bf Avg. LQ} & \bf{TPs} & {\bf FPs} & {\bf FNs} \\
\hline
\hline
0.05 & 22.084 & 0.399 & 0.835 & 0.445 & 0.497 & \textbf{59026} & 112799 & \textbf{29063} \\
\hline
\textbf{0.1} & \textbf{22.569} & \textbf{0.409} & \textbf{0.835} & \textbf{0.464} & 0.498 & 58961 & 112689 & 29128  \\
\hline
0.2 & 19.861 & 0.360 & 0.835 & 0.383 & 0.499 & 58848 & \textbf{112414} & 29241 \\
\hline
0.3 & 12.882 & 0.235 & 0.834 & 0.219 & \textbf{0.500} & 58509 & 112471 & 29580 \\
\hline
\end{tabular}
\caption{The results of an ensemble model using different bounding box ratios on
the validation dataset (Avg. SQ=average spatial quality, Avg. LQ=average label
quality, TPs=true positives, FPs=false positives, FNs=false negatives).}
\label{tab:validation_set_results_2_boxratio}
\end{table*}

\begin{table*}
\centering
\begin{tabular}{|c|c|c|c|c|c|c|c|c|}
\hline
{\bf Covariance Scale} & {\bf PDQ Score} & {\bf Avg. pPDQ} & {\bf Avg. FP} & {\bf Avg. SQ} & {\bf Avg. LQ} & \bf{TPs} & {\bf FPs} & {\bf FNs} \\
\hline
\hline
0.1 & 21.894 & 0.402 & 0.834 & 0.462 & \textbf{0.502} & 58275 & 113517 & 29814 \\
\hline
0.2 & 22.536 & \textbf{0.411} & \textbf{0.835} & \textbf{0.473} & 0.500 & 58605 & 113146 & 29484 \\
\hline
\textbf{0.3} & \textbf{22.569} & 0.409 & 0.835 & 0.464 & 0.498 & 58961 & 112689 & 29128  \\
\hline
0.4 & 22.458 & 0.403 & 0.835 & 0.450 & 0.495 & 59421 & 112020 & 28668 \\
\hline
0.5 & 22.283 & 0.397 & 0.835 & 0.437 & 0.492 & \textbf{59737} & \textbf{111557} & \textbf{28352} \\
\hline
\end{tabular}
\caption{The results of an ensemble model using different covariance scales on
the validation dataset (Avg. SQ=average spatial quality, Avg. LQ=average label
quality, TPs=true positives, FPs=false positives, FNs=false negatives).}
\label{tab:validation_set_results_3_covariance}
\end{table*}

\begin{table*}
\centering
\begin{tabular}{|c|c|c|c|c|c|c|c|c|}
\hline
{\bf IoU Threshold} & {\bf PDQ Score} & {\bf Avg. pPDQ} & {\bf Avg. FP} & {\bf Avg. SQ} & {\bf Avg. LQ} & \bf{TPs} & {\bf FPs} & {\bf FNs} \\
\hline
\hline
0.1 & 21.966 & \textbf{0.419} & 0.817 & 0.468 & \textbf{0.514} & 54194 & \textbf{84036} & 33895 \\
\hline
0.2 & 22.395 & 0.411 & 0.827 & 0.465 & 0.504 & 57398 & 100192 & 30691 \\
\hline
\textbf{0.3} & \textbf{22.569} & 0.409 & 0.835 & 0.464 & 0.498 & 58961 & 112689 & 29128  \\
\hline
0.4 & 22.530 & 0.405 & 0.844 & 0.465 & 0.486 & 60824 & 136205 & 27265 \\
\hline
0.5 & 22.314 & 0.405 & \textbf{0.856} & \textbf{0.472} & 0.477 & \textbf{62338} & 175034 & \textbf{25751} \\
\hline
\end{tabular}
\caption{The results of an ensemble model using different IoU threshold values
on the validation set (Avg. SQ=average spatial quality, Avg. LQ=average label
quality, TPs=true positives, FPs=false positives, FNs=false negatives).}
\label{tab:validation_set_results_4_iouthreshold}
\end{table*}

\begin{table*}
\centering
\begin{tabular}{|c|c|c|c|c|c|c|c|c|}
\hline
{\bf Method} & {\bf PDQ Score} & {\bf Avg. pPDQ}  & {\bf Avg. FP} & {\bf Avg. SQ} & {\bf Avg. LQ} & \bf{TPs} & {\bf FPs} & {\bf FNs} \\
\hline
\hline
HTC & 22.769 & 0.415 & 0.833 & \textbf{0.472} & 0.496 & 56054 & 84565 & 32035 \\
\hline
\textbf{HTC + MC dropout} & \textbf{22.822} & 0.414 & 0.824 & 0.467 & 0.504 & 58498 & 102610 & 29591\\
\hline
Grid R-CNN & 18.014 & 0.410 & 0.799 & 0.465 & 0.514 & 42383 & \textbf{42162} & 45706 \\
\hline
Grid R-CNN + MC dropout & 18.222 & 0.395 & 0.789 & 0.445 & 0.514 & 48340 & 79423 & 39749 \\
\hline
Ensemble & 22.569 & 0.409 & \textbf{0.835} & 0.464 & 0.498 & 58961 & 112689 & 29128 \\
\hline
\textbf{Ensemble + MC dropout} & \textbf{22.599} & \textbf{0.417} & 0.818 & 0.469 & \textbf{0.515} & \textbf{59847} & 122280 & \textbf{28242} \\
\hline
\end{tabular}
\caption{The results of a single and an ensemble model before and after adding dropout
on the validation dataset (Avg. SQ=average spatial quality, Avg. LQ=average
label quality, TPs=true positives, FPs=false positives, FNs=false
negatives).}
\label{tab:validation_set_results_5}
\end{table*}

\begin{table*}
\centering
\begin{tabular}{|c|c|c|c|c|c|c|c|c|}
\hline
{\bf Method} & {\bf PDQ Score} & {\bf Avg. pPDQ}  & {\bf Avg. FP} & {\bf Avg. SQ} & {\bf Avg. LQ} & \bf{TPs} & {\bf FPs} & {\bf FNs} \\
\hline
\hline
Ensemble (averaging) & 12.642 & 0.223 & \bf{0.907} & 0.324 & 0.238 & 54622 &
\bf{90024} & 33467 \\
\hline
Ensemble (same labels, averaging) & 14.286 & 0.246 & 0.887 & 0.323 & 0.294 & 57425 & 101779 & 30664 \\
\hline
Ensemble (ours) & \bf{22.569} & \bf{0.409} & 0.835 & \bf{0.464} & \bf{0.498}
& \bf{58961} & 112689 & \bf{29128} \\
\hline
\end{tabular}
\caption{A comparison of ensemble models using averaging and our merging 
strategy on the validation dataset (Avg. SQ=average spatial quality, Avg. 
LQ=average label quality, TPs=true positives, FPs=false positives, FNs=false
negatives).}
\label{tab:validation_set_results_6}
\end{table*}

\section{Conclusion and Future Work}
\label{sec:conclusion_and_future_work}
We presented an uncertainty estimation framework which makes use of the recently
introduced PDQ evaluation metric and achieves a top score on the PrOD challenge
validation dataset. By extensively exploring the hyperparameter space, we showed
the quality of object detections can be maintained by introducing additional
post-processing techniques. Likewise, we've shown that post-processing methods
designed to increase detection quality may influence one another and the
optimization of these techniques can be done offline. We highlighted that
the IoU parameter of NMS designed for mAP does not lead to high-quality
detections. In fact, we experimentally demonstrated that very low IoU
thresholds group more detections together and reduce the number of false
positives.

Our forthcoming work in probabilistic object detection will explore the
incorporation of manual heuristics into an end-to-end model. For example, it is
advantageous to add bias for smaller bounding boxes. Moreover, post-processing
steps must be adaptive, i.e., the hyperparameters for these steps may be image
dependent. This warrants further investigation into the adequacy of applying
computationally intensive methods that occasionally decrease the quality of the
performance. More importantly, when applying these methods one should clearly
select detectors based on their strengths and weaknesses while being aware of
any potential increase in computation.


\bibliographystyle{IEEEtran}
\bibliography{IEEEabrv,an_uncertainty_estimation_framework_for_probabilistic_object_detection}

\begin{thebibliography}{10}
\providecommand{\url}[1]{#1}
\csname url@samestyle\endcsname
\providecommand{\newblock}{\relax}
\providecommand{\bibinfo}[2]{#2}
\providecommand{\BIBentrySTDinterwordspacing}{\spaceskip=0pt\relax}
\providecommand{\BIBentryALTinterwordstretchfactor}{4}
\providecommand{\BIBentryALTinterwordspacing}{\spaceskip=\fontdimen2\font plus
\BIBentryALTinterwordstretchfactor\fontdimen3\font minus
  \fontdimen4\font\relax}
\providecommand{\BIBforeignlanguage}[2]{{%
\expandafter\ifx\csname l@#1\endcsname\relax
\typeout{** WARNING: IEEEtran.bst: No hyphenation pattern has been}%
\typeout{** loaded for the language `#1'. Using the pattern for}%
\typeout{** the default language instead.}%
\else
\language=\csname l@#1\endcsname
\fi
#2}}
\providecommand{\BIBdecl}{\relax}
\BIBdecl

\bibitem{russakovsky2015imagenet}
O.~Russakovsky, J.~Deng, H.~Su, J.~Krause, S.~Satheesh, S.~Ma, Z.~Huang,
  A.~Karpathy, A.~Khosla, M.~Bernstein \emph{et~al.}, ``Imagenet large scale
  visual recognition challenge,'' \emph{International Journal of Computer
  Vision}, vol. 115, no.~3, pp. 211--252, 2015.

\bibitem{everingham2010pascal}
M.~Everingham, L.~Van~Gool, C.~K. Williams, J.~Winn, and A.~Zisserman, ``The
  pascal visual object classes (voc) challenge,'' \emph{International Journal
  of Computer Vision}, vol.~88, no.~2, pp. 303--338, 2010.

\bibitem{everingham2015pascal}
M.~Everingham, S.~A. Eslami, L.~Van~Gool, C.~K. Williams, J.~Winn, and
  A.~Zisserman, ``The pascal visual object classes challenge: A
  retrospective,'' \emph{International Journal of Computer Vision}, vol. 111,
  no.~1, pp. 98--136, 2015.

\bibitem{lin2014microsoft}
T.-Y. Lin, M.~Maire, S.~Belongie, J.~Hays, P.~Perona, D.~Ramanan,
  P.~Doll{\'a}r, and C.~L. Zitnick, ``Microsoft coco: Common objects in
  context,'' in \emph{Proceedings of the European Conference on Computer Vision
  (ECCV)}.\hskip 1em plus 0.5em minus 0.4em\relax Springer, 2014, pp. 740--755.

\bibitem{skinner2019probabilistic}
J.~Skinner, D.~Hall, H.~Zhang, F.~Dayoub, and N.~S{\"u}nderhauf, ``The
  probabilistic object detection challenge,'' \emph{arXiv preprint
  arXiv:1903.07840}, 2019.

\bibitem{hall2020probabilistic}
D.~Hall, F.~Dayoub, J.~Skinner, H.~Zhang, D.~Miller, P.~Corke, G.~Carneiro,
  A.~Angelova, and N.~S{\"u}nderhauf, ``Probabilistic object detection:
  definition and evaluation,'' in \emph{Proceedings of the IEEE Winter
  Conference on Applications of Computer Vision (WACV)}, 2020, pp. 1031--1040.

\bibitem{lyu2020probabilistic}
Z.~Lyu, N.~Gutierrez, A.~Rajguru, and W.~J. Beksi, ``Probabilistic object
  detection via deep ensembles,'' in \emph{Proceedings of the European
  Conference on Computer Vision (ECCV) Workshops}.\hskip 1em plus 0.5em minus
  0.4em\relax Springer, 2020, pp. 67--75.

\bibitem{lakshminarayanan2017simple}
B.~Lakshminarayanan, A.~Pritzel, and C.~Blundell, ``Simple and scalable
  predictive uncertainty estimation using deep ensembles,'' in
  \emph{Proceedings of the Advances in Neural Information Processing Systems
  (NeurIPS)}, 2017, pp. 6402--6413.

\bibitem{gal2016dropout}
Y.~Gal and Z.~Ghahramani, ``Dropout as a bayesian approximation: Representing
  model uncertainty in deep learning,'' in \emph{Proceedings of the
  International Conference on Machine Learning (ICML)}, 2016, pp. 1050--1059.

\bibitem{chen2019hybrid}
K.~Chen, J.~Pang, J.~Wang, Y.~Xiong, X.~Li, S.~Sun, W.~Feng, Z.~Liu, J.~Shi,
  W.~Ouyang \emph{et~al.}, ``Hybrid task cascade for instance segmentation,''
  in \emph{Proceedings of the IEEE Conference on Computer Vision and Pattern
  Recognition (CVPR)}, 2019, pp. 4974--4983.

\bibitem{lu2019grid}
X.~Lu, B.~Li, Y.~Yue, Q.~Li, and J.~Yan, ``Grid r-cnn,'' in \emph{Proceedings
  of the IEEE Conference on Computer Vision and Pattern Recognition (CVPR)},
  2019, pp. 7363--7372.

\bibitem{depod}
\url{https://github.com/robotic-vision-lab/Deep-Ensembles-For-Probabilistic-Object-Detection}.

\bibitem{szegedy2013deep}
C.~Szegedy, A.~Toshev, and D.~Erhan, ``Deep neural networks for object
  detection,'' in \emph{Proceedings of the Advances in Neural Information
  Processing Systems (NeurIPS)}, 2013, pp. 2553--2561.

\bibitem{redmon2016you}
J.~Redmon, S.~Divvala, R.~Girshick, and A.~Farhadi, ``You only look once:
  Unified, real-time object detection,'' in \emph{Proceedings of the IEEE
  Conference on Computer Vision and Pattern Recognition (CVPR)}, 2016, pp.
  779--788.

\bibitem{liu2016ssd}
W.~Liu, D.~Anguelov, D.~Erhan, C.~Szegedy, S.~Reed, C.-Y. Fu, and A.~C. Berg,
  ``Ssd: Single shot multibox detector,'' in \emph{Proceedings of the European
  Conference on Computer Vision (ECCV)}.\hskip 1em plus 0.5em minus 0.4em\relax
  Springer, 2016, pp. 21--37.

\bibitem{redmon2017yolo9000}
J.~Redmon and A.~Farhadi, ``Yolo9000: better, faster, stronger,'' in
  \emph{Proceedings of the IEEE Conference on Computer Vision and Pattern
  Recognition (CVPR)}, 2017, pp. 7263--7271.

\bibitem{redmon2018yolov3}
------, ``Yolov3: An incremental improvement,'' \emph{arXiv preprint
  arXiv:1804.02767}, 2018.

\bibitem{bochkovskiy2020yolov4}
A.~Bochkovskiy, C.-Y. Wang, and H.-Y.~M. Liao, ``Yolov4: Optimal speed and
  accuracy of object detection,'' \emph{arXiv preprint arXiv:2004.10934}, 2020.

\bibitem{girshick2014rich}
R.~Girshick, J.~Donahue, T.~Darrell, and J.~Malik, ``Rich feature hierarchies
  for accurate object detection and semantic segmentation,'' in
  \emph{Proceedings of the IEEE Conference on Computer Vision and Pattern
  Recognition (CVPR)}, 2014, pp. 580--587.

\bibitem{girshick2015fast}
R.~Girshick, ``Fast r-cnn,'' in \emph{Proceedings of the IEEE International
  Conference on Computer Vision (ICCV)}, 2015, pp. 1440--1448.

\bibitem{ren2015faster}
S.~Ren, K.~He, R.~Girshick, and J.~Sun, ``Faster r-cnn: Towards real-time
  object detection with region proposal networks,'' in \emph{Proceedings of the
  Advances in Neural Information Processing Systems (NeurIPS)}, 2015, pp.
  91--99.

\bibitem{he2017mask}
K.~He, G.~Gkioxari, P.~Doll{\'a}r, and R.~Girshick, ``Mask r-cnn,'' in
  \emph{Proceedings of the IEEE Conference on Computer Vision and Pattern
  Recognition (CVPR)}, 2017, pp. 2961--2969.

\bibitem{cai2018cascade}
Z.~Cai and N.~Vasconcelos, ``Cascade r-cnn: Delving into high quality object
  detection,'' in \emph{Proceedings of the IEEE Conference on Computer Vision
  and Pattern Recognition (CVPR)}, 2018, pp. 6154--6162.

\bibitem{denker1991transforming}
J.~S. Denker and Y.~LeCun, ``Transforming neural-net output levels to
  probability distributions,'' in \emph{Proceedings of the Advances in Neural
  Information Processing Systems (NeurIPS)}, 1991, pp. 853--859.

\bibitem{mackay1992practical}
D.~J. MacKay, ``A practical bayesian framework for backpropagation networks,''
  \emph{Neural Computation}, vol.~4, no.~3, pp. 448--472, 1992.

\bibitem{hernandez2015probabilistic}
J.~M. Hern{\'a}ndez-Lobato and R.~Adams, ``Probabilistic backpropagation for
  scalable learning of bayesian neural networks,'' in \emph{Proceedings of the
  International Conference on Machine Learning (ICML)}, 2015, pp. 1861--1869.

\bibitem{hinton2012improving}
G.~E. Hinton, N.~Srivastava, A.~Krizhevsky, I.~Sutskever, and R.~R.
  Salakhutdinov, ``Improving neural networks by preventing co-adaptation of
  feature detectors,'' \emph{arXiv preprint arXiv:1207.0580}, 2012.

\bibitem{srivastava2014dropout}
N.~Srivastava, G.~Hinton, A.~Krizhevsky, I.~Sutskever, and R.~Salakhutdinov,
  ``Dropout: a simple way to prevent neural networks from overfitting,''
  \emph{Journal of Machine Learning Research}, vol.~15, no.~1, pp. 1929--1958,
  2014.

\bibitem{maeda2014bayesian}
S.~Maeda, ``A bayesian encourages dropout,'' \emph{arXiv preprint
  arXiv:1412.7003}, 2014.

\bibitem{gal2015bayesian}
Y.~Gal and Z.~Ghahramani, ``Bayesian convolutional neural networks with
  bernoulli approximate variational inference,'' \emph{arXiv preprint
  arXiv:1506.02158}, 2015.

\bibitem{miller2018dropout}
D.~Miller, L.~Nicholson, F.~Dayoub, and N.~S{\"u}nderhauf, ``Dropout sampling
  for robust object detection in open-set conditions,'' in \emph{Proceedings of
  the IEEE International Conference on Robotics and Automation (ICRA)}, 2018,
  pp. 1--7.

\bibitem{miller2019evaluating}
D.~Miller, F.~Dayoub, M.~Milford, and N.~S{\"u}nderhauf, ``Evaluating merging
  strategies for sampling-based uncertainty techniques in object detection,''
  in \emph{Proceedings of the IEEE International Conference on Robotics and
  Automation (ICRA)}, 2019, pp. 2348--2354.

\bibitem{kraus2019uncertainty}
F.~Kraus and K.~Dietmayer, ``Uncertainty estimation in one-stage object
  detection,'' in \emph{Proceedings of the IEEE Intelligent Transportation
  Systems Conference (ITSC)}, 2019, pp. 53--60.

\bibitem{kendall2017multitask}
A.~Kendall, Y.~Gal, and R.~Cipolla, ``Multi-task learning using uncertainty to
  weigh losses for scene geometry and semantics,'' \emph{arXiv preprint
  arXiv:1705.07115}, 2017.

\bibitem{sirinukunwattana2016locality}
K.~Sirinukunwattana, S.~E.~A. Raza, Y.-W. Tsang, D.~R. Snead, I.~A. Cree, and
  N.~M. Rajpoot, ``Locality sensitive deep learning for detection and
  classification of nuclei in routine colon cancer histology images,''
  \emph{IEEE Transactions on Medical Imaging}, vol.~35, no.~5, pp. 1196--1206,
  2016.

\bibitem{tanno2017bayesian}
R.~Tanno, D.~E. Worrall, A.~Ghosh, E.~Kaden, S.~N. Sotiropoulos, A.~Criminisi,
  and D.~C. Alexander, ``Bayesian image quality transfer with cnns: exploring
  uncertainty in dmri super-resolution,'' in \emph{Proceedings of the
  International Conference on Medical Image Computing and Computer-Assisted
  Intervention}.\hskip 1em plus 0.5em minus 0.4em\relax Springer, 2017, pp.
  611--619.

\bibitem{choi2019aaussianya}
J.~Choi, D.~Chun, H.~Kim, and H.~Lee, ``Gaussian yolov3: An accurate and fast
  object detector using localization uncertainty for autonomous driving,''
  \emph{Proceedings of the IEEE International Conference on Computer Vision
  (ICCV)}, pp. 502--511, 2019.

\bibitem{wang2019augpod}
C.-W. Wang, C.-A. Cheng, C.-J. Cheng, H.-N. Hu, H.-K. Chu, and M.~Sun,
  ``Augpod: Augmentation-oriented probabilistic object detection,'' in
  \emph{Proceedings of the CVPR Workshop on the Robotic Vision Probabilistic
  Object Detection Challenge}, 2019.

\bibitem{ammirato2019mask}
P.~Ammirato and A.~C. Berg, ``A mask-rcnn baseline for probabilistic object
  detection,'' in \emph{Proceedings of the CVPR Workshop on the Robotic Vision
  Probabilistic Object Detection Challenge}, 2019.

\bibitem{li2019teamgl}
D.~Li, C.~Xu, Y.~Liu, and Z.~Qin, ``Teamgl at acrv robotic vision challenge 1:
  Probabilistic object detection via staged non-suppression ensembling,'' in
  \emph{Proceedings of the CVPR Workshop on the Robotic Vision Probabilistic
  Object Detection Challenge}, 2019.

\bibitem{chen2019mmdetection}
K.~Chen, J.~Wang, J.~Pang, Y.~Cao, Y.~Xiong, X.~Li, S.~Sun, W.~Feng, Z.~Liu,
  J.~Xu \emph{et~al.}, ``Mmdetection: Open mmlab detection toolbox and
  benchmark,'' \emph{arXiv preprint arXiv:1906.07155}, 2019.

\end{thebibliography}
\end{document}